\documentclass[conference]{IEEEtran}
\IEEEoverridecommandlockouts
\usepackage{cite}
\usepackage{amsmath,amssymb,amsfonts}
\usepackage{algorithmic}
\usepackage{graphicx}
\usepackage{textcomp}
\usepackage{xcolor}
\usepackage{booktabs}
\usepackage{graphicx}
\usepackage{multirow}
\usepackage{threeparttable}    
\usepackage{booktabs}
\usepackage{diagbox} 
\usepackage{hyperref}
\def\BibTeX{{\rm B\kern-.05em{\sc i\kern-.025em b}\kern-.08em
    T\kern-.1667em\lower.7ex\hbox{E}\kern-.125emX}}
\begin{document}

\title{CLIP-DQA: Blindly Evaluating Dehazed Images from Global and Local Perspectives Using CLIP 
\author{\IEEEauthorblockN{Yirui	Zeng$^{1,\ast}$, Jun Fu$^{1,\ast}$, Hadi Amirpour$^{2}$, Huasheng Wang$^{3}$, Guanghui Yue$^{4}$, Hantao Liu$^{1}$, Ying Chen$^{3}$, Wei Zhou$^{1,\dagger}$ \\
$^{1}$School of Computer Science and Informatics, Cardiff University, United Kingdom \\
$^{2}$Christian Doppler Laboratory ATHENA, Alpen-Adria-Universit$\ddot{a}$t, Klagenfurt, Austria \\
$^{3}$Alibaba Group \\
$^{4}$School of Biomedical Engineering, Shenzhen University, China \\
Email: zengyr5@mail2.sysu.edu.cn \qquad fujun@mail.ustc.edu.cn \qquad zhouw26@cardiff.ac.uk}}
\thanks{ $\ast$ denotes equal contribution, and $\dagger$ indicates the corresponding author.}
}


\maketitle

\begin{abstract}
Blind dehazed image quality assessment (BDQA), which aims to accurately predict the visual quality of dehazed images without any reference information, is essential for the evaluation, comparison, and optimization of image dehazing algorithms. Existing learning-based BDQA methods have achieved remarkable success, while the small scale of DQA datasets limits their performance. To address this issue, in this paper, we propose to adapt Contrastive Language-Image Pre-Training (CLIP), pre-trained on large-scale image-text pairs, to the BDQA task. Specifically, inspired by the fact that the human visual system understands images based on hierarchical features, we take global and local information of the dehazed image as the input of CLIP. To accurately map the input hierarchical information of dehazed images into the quality score, we tune both the vision branch and language branch of CLIP with prompt learning. Experimental results on two authentic DQA datasets demonstrate that our proposed approach, named CLIP-DQA, achieves more accurate quality predictions over existing BDQA methods. The code is available at \url{https://github.com/JunFu1995/CLIP-DQA}.

\end{abstract}


\section{Introduction}
Haze is a common natural phenomenon that significantly reduces visibility in scenes, causing many computer vision algorithms, such as object detection~\cite{chen2021disentangle,xie2023mutual} and image recognition~\cite{wang2020category}, to experience severe performance degradation.  To alleviate this issue, considerable image dehazing algorithms (DHAs)~\cite{he2010single,cai2016dehazenet,liu2021joint,li2022image,song2023vision,cheng2024progressive,wang2024hazeclip} have been proposed. However, before deploying these DHAs at scale, it is necessary to evaluate their effectiveness, \textit{i.e.,} assess the quality of dehazed images they generate. 

The most accurate way to measure the quality of dehazed images is subjective quality evaluation~\cite{zhou2024dehazed}, where the quality of dehazed images is directly evaluated by a certain number of human subjects. However, subjective quality evaluation has a narrow range of applications since it needs to conduct time-consuming and labor-intensive subjective experiments. As a result, objective quality evaluation is proposed, aiming to automatically assess the quality of dehazed images without human involvement. 

In general, objective dehazed quality measures can be divided into three categories: full-reference dehazed image quality assessment~\cite{min2019quality,zhao2020dehazing,liu2020image}, reduced-reference dehazed image quality assessment~\cite{fang2011image,song2017dehazed,min2018objective,wang2019pixel,zeng2023trg,guan2023dual}, and no-reference dehazed image quality assessment~\cite{choi2015referenceless,shen2017blind,guan2022visibility,zhou2024dehazed,lv2023blind}. When evaluating the quality of dehazed images, the former two categories require reference images, while the last category only takes dehazed images as input. In real-world scenarios, reference images are typically unavailable. Therefore, NR DQA, also known as blind dehazed image quality assessment (BDQA), has received considerable attention in recent years.

Existing BDQA approaches are mainly composed of traditional methods~\cite{choi2015referenceless,shen2017blind,guan2022visibility,zhou2024dehazed} and learning-based methods~\cite{zhang2020hazdesnet,lv2023blind,zeng2023trg}. Traditional methods usually manually design some haze-related features for quality evaluation. Since handcrafted features have limited capabilities in representing distortion and content of dehazed images, the performance of traditional BDQA methods is often unsatisfactory. To this end, learning-based methods employ deep neural networks to automatically extract representative features from dehazed images. Compared to traditional BDQA methods, learning-based BDQA methods achieve more accurate quality prediction. However, their performance is still limited by the small size of the DQA dataset.

To alleviate this issue, we resort to Contrastive Language-Image Pre-Training (CLIP)~\cite{radford2021learning}, which has shown good generalization ability across various image recognition tasks~\cite{li2024graphadapter,zhang2022pointclip}. In the BDQA task, the perceptual characteristics of the human visual system need to be considered~\cite{zhou2024dehazed}, that is, image understanding based on hierarchical features. Therefore, we feed patches that preserve the local details of the dehazed image and the resized image that preserves the global structure of the dehazed image into CLIP. Then, we tune CLIP by layerwisely inserting learnable prompts into its vision and language branches to accurately map the input hierarchical information of dehazed images into the quality score.

The contributions of this paper are mainly two-fold:
\begin{itemize}
	\item We present the first CLIP-based BDQA method, CLIP-DQA, which evaluates dehazed images from global and local perspectives. 
	\item We conduct extensive experiments, including ablation studies and visualizations, to verify the efficacy of the proposed method.
\end{itemize}

\section{Proposed Method}
\begin{figure*}[t]
	\centering
	\includegraphics[width=0.85\linewidth]{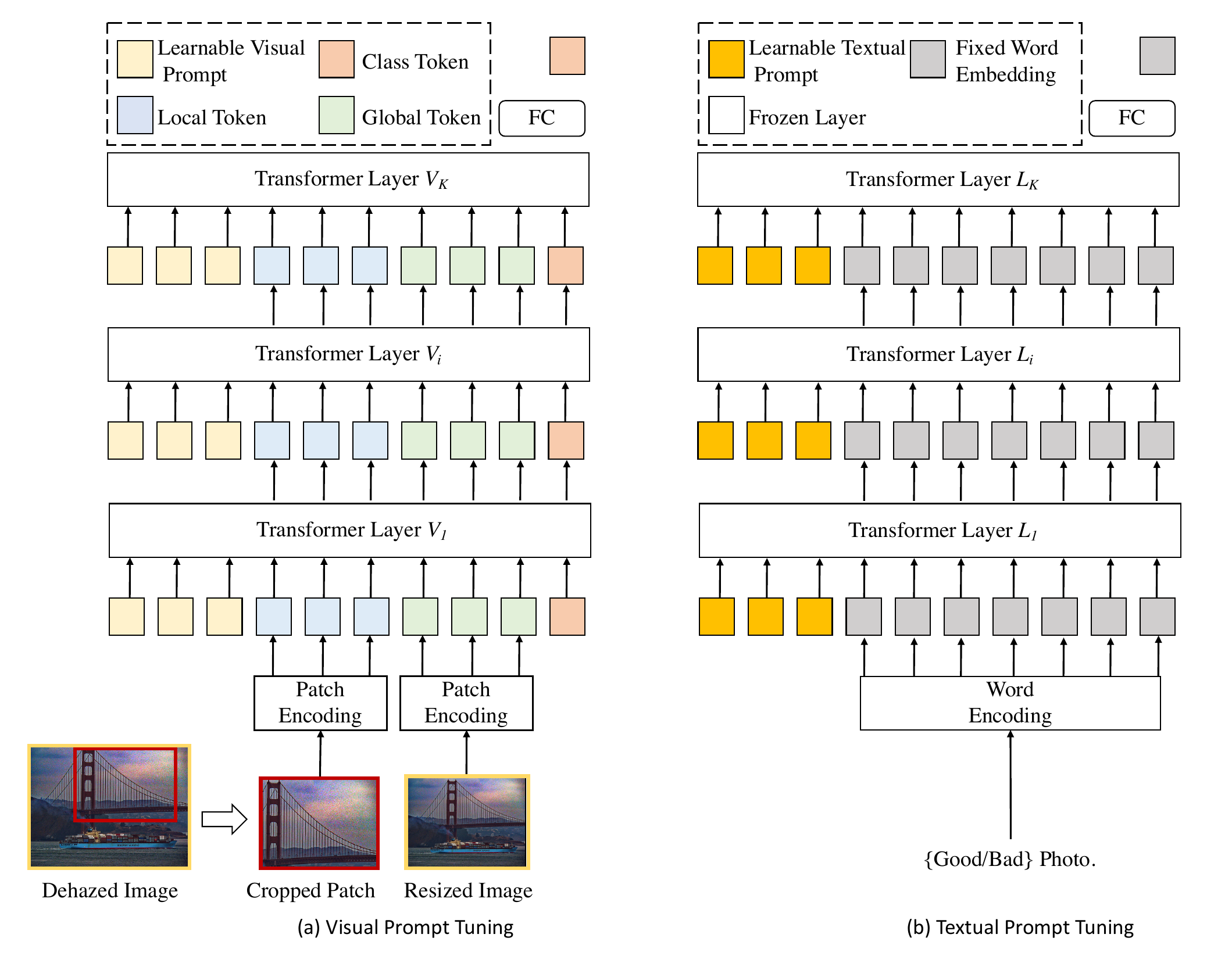}
	\caption{Illustration of multi-modal prompt tuning for blind dehazed image quality assessment. ``FC'' means the fully connected layer. We use ``\texttt{Good photo.}'' and ''\texttt{Bad photo.}'' as antonym text prompts following the CLIP-based general image quality assessment method~\cite{wang2023exploring}. }
	\label{fig:fr}
\end{figure*}

\subsection{Problem Formulation}
Given a dehazed image $I$, blind dehazed image quality assessment aims to estimate the visual quality score of the dehazed image without any reference information. In existing works, a patch-based evaluation framework is widely used for BDQA. In such a framework, we first crop $N$ patches $\{I^i_p\}^{i=N-1}_{i=0}$ from the input dehazed image $I$, then estimate the quality score for each patch, and finally use the average score as the quality score of the whole picture:
\begin{equation}
\hat{Q} = \frac{1}{N} \sum_{i=0}^{N-1} f(I^i_p),
\end{equation} 
where $f(\cdot)$ denotes the BDQA method, and $\hat{Q}$ is the estimated result.

Inspired by the human visual system using hierarchical features for image understanding, we evaluate the dehazed image from global and local perspectives:
\begin{equation}
\hat{Q} = \frac{1}{N} \sum_{i=0}^{N-1} f(I^i_p, I_s),
\label{eq:pf}
\end{equation} 
where $I^i_p$ keeps the local details and $I_s$, a resized version of $I$, maintains the global structure.  Moreover, we model $f(\cdot)$ using CLIP, which learns well-generalized knowledge from millions of image-text pairs. Next, we will detail how to evaluate dehazed images with CLIP.

\subsection{Zero-shot BDQA}
Since CLIP can judge the similarity between images and natural language descriptions, we can directly use CLIP to evaluate the dehazed image without training, \textit{i.e.,} zero-shot BDQA. 

Let $T_p$ and $T_n$ be a pair of antonym text prompts, \textit{e.g.,} ``\texttt{Good photo.}'' and ''\texttt{Bad photo.}'' . The estimated visual quality, $f(I^i_p, I_s)$ in Equation.~\ref{eq:pf}, can be calculated as follows:
\begin{equation}
f(I^i_p, I_s) = \frac{e^{sim(t_p, c_i)}}{e^{sim(tp, c_i)}+e^{sim(t_n, c_i)}},
\end{equation} 
where $t_p$ and $t_s$ are the textual representation extracted by the language branch of CLIP from $T_p$ and $T_n$, respectively. $c_i$ is the visual representation extracted by the vision branch of CLIP from the pair of $I^i_p$ and $I_s$. Here, $sim(\cdot, \cdot)$ calculates the cosine distance between textual and visual representations. 

However, the performance of the zero-shot BDQA is typically far from satisfactory. There are two main reasons for this result. First, the handcrafted antonym text prompts are often sub-optimal for BDQA since designing effective antonym text prompts requires considerable expertise. Second, the visual representation
extracted by CLIP may not be discriminative for BDQA as dehazed images differ from images used for training CLIP in terms of distortion type and appearance. 

\subsection{Multi-modal Prompt Tuning}
In order to better adapt CLIP to BDQA, it is necessary to fine-tune CLIP using the DQA dataset. Motivated by the success of prompt learning~\cite{zhou2022conditional,jia2022visual,zhou2022learning,khattak2023maple}, we tune CLIP with multi-modal prompts, including textual and visual prompt tuning, as shown in Fig.~\ref{fig:fr}. The core idea of multi-modal prompt tuning is to use learnable textual prompts for automatically mining useful soft antonym text prompts from DQA datasets, and use learnable visual prompts for mitigating the domain gap between dehazed images and natural images. 

\noindent\textbf{Textual Prompt Tuning.} The language branch of CLIP consists of $K$ transformer layers, and the $i$-th transformer layer can be defined as:
\begin{equation}
[{W}_i] = \mathcal{L}_{i}({W}_{i-1}), \quad i = 1, \cdots, K,
\end{equation}
where ${W}_{i-1}$ and ${W}_i$ are the input and output of the $i$-th transformer layer $\mathcal{L}_{i}$, respectively. The input of the first transform layer, $W_0$, corresponds to word embeddings of handcrafted antonym text prompts. To tune the language branch of CLIP, we layerwisely insert learnable prompts:
\begin{equation}
[\ \underline{\hspace{0.3cm}}\ , \ W_{i}] = \mathcal{L}_{i}([F_{i-1}(P_{i-1}), W_{i-1}]),
\label{eq:tp}
\end{equation}
where we map the set of learnable prompts $P_{i-1}$ into the same space as $W_{i-1}$ through a fully-connected layer $F_{i-1}$. Notably, the output of $P_{i-1}$ is discarded after the $i$-th transformer layer $L_i$.

\noindent\textbf{Visual Prompt Tuning.} In this paper, the vision branch of CLIP is transformer-based and also contains $K$ transformer layers. The formulation of each transformer layer is defined as follows:
\begin{equation}
[c_i, E_i^{l}, E_i^{g} \ ] = \mathcal{V}_{i}([c_{i-1}, E_{i-1}^{l}, E_{i-1}^{g}), \quad i=1, \cdots, K,
\label{eq:vpp}
\end{equation}
where $c_i$, $E_i^{l}$, and $E_i^{g}$ are the class token, the token set of the input patch, and the token set of the input resized dehazed image, respectively. To tune the vision branch of CLIP, we also layerwisely insert learnable prompts:
\begin{equation}
[c_i, E_i^{l}, E_i^{g}, \ \underline{\hspace{0.3cm}} \ ] = \mathcal{V}_{i}([c_{i-1}, E_{i-1}^{l}, E_{i-1}^{g}, \hat{F}_{i-1}(\hat{P}_{i-1})]),
\label{eq:vpp}
\end{equation}
where we map the set of learnable prompts $\hat{P}_{i-1}$ into the same space as $E_{i-1}$ through a fully-connected layer $\hat{F}_{i-1}$. Similar to textual prompt tuning, we discard the output of the learnable prompts $\hat{P}_{i-1}$ after the transformer layer $V_i$.

\subsection{Loss Function}
For the DQA task, we use Mean Square Error (MSE) as the training objective:
\begin{equation}
L_{MSE} = \frac{1}{B} \sum_{j=1}^{B}\lVert \hat{Q}_j - Q_j\rVert^2_2,
\end{equation}
where $B$ is the batch size, $\hat{Q}_j$ and $Q_j$ are the predicted quality score and the mean opinion score of the $j$-th dehazed image.

\section{Experiments}
\subsection{Experimental Protocols}
To verify the effectiveness of our proposed method, we conduct experiments on two authentic dehazed image quality databases:
\begin{itemize}
    \item DHQ database~\cite{min2018objective}: It consists of 250 hazy images and 1,750 dehazed images generated by 7 image dehazing algorithms. Each dehazed image is labeled by a mean opinion score (MOS) ranging from 0 to 100.  
    \item exBeDDE database~\cite{zhao2020dehazing}: It contains 12 haze-free images, 167 hazy images, and 1,670 dehazed images produced by 10 image dehazing algorithms. Each dehazed image is annotated by a MOS ranging from 0 to 1.  
\end{itemize}

As suggested by the video quality expert group (VQEG)~\cite{antkowiak2000final}, we employ Spearman rank order correlation coefficient (SRCC), Pearson linear correlation coefficient (PLCC), and Kendall rank order correlation coefficient (KRCC) for performance comparisons. All three evaluation criteria ranged from 0 to 1, and the higher the value, the better the performance. 

For fair comparison, we evaluate each image quality assessment method 10 times, and report the average results. At each time, we randomly split the dataset into two parts based on the content of the image, 80\% for training and 20\% for testing.

We build the proposed method based on ViT-B/32 CLIP, and the length of learnable prompts at each transformer layer is set to 8. During training, we keep CLIP frozen and optimize the remaining parts using the Adam optimizer ~\cite{adam}  with a learning rate of 1e-4. The total training epoch and the batch size are 50 and 64, respectively. All experiments are run on a single NVIDIA RTX 4090 GPU.

\begin{table}[t]
	\caption{The results of different methods on the DHQ and exBeDDE datasets.}
	\centering
	\scalebox{0.75}{
            \begin{threeparttable}          
		\begin{tabular}{c| c| c cc| ccc}
			\toprule
			\multirow{2}{*}{Types}  & \multirow{2}{*}{Method} & \multicolumn{3}{c|}{DHQ} & \multicolumn{3}{c}{exBeDDE} \\  \cline{3-8}
			& & SRCC & PLCC & KRCC & SRCC & PLCC & KRCC  \\\midrule
			\multirow{3}{*}{\shortstack{ FR \\ GIQA}}& PSNR & - & - & - & 0.5375 & 0.5129 & 0.3758 \\
			& SSIM~\cite{wang2004image} &  - & - & - & 0.5730 & 0.5584 & 0.3990 \\
			& VSI~\cite{zhang2014vsi} &  - & - & - & 0.5895 & 0.5789 & 0.4145  \\ \midrule     
			\multirow{2}{*}{\shortstack{ FR \\ DQA}} & VI~\cite{zhao2020dehazing} &  - & - & - & 0.5089 & 0.4909 & 0.3545 \\
			& RI~\cite{zhao2020dehazing} &  - & - & - &  0.5013 & 0.4969 & 0.3515 \\ \midrule      
			\multirow{3}{*}{\shortstack{ NR \\ GIQA}}& CNNIQA~\cite{cnniqa}& 0.6756&0.6937 & 0.4911& 0.8435 & 0.8142 & 0.6482\\
			& TReS~\cite{tres}  & 0.8395 & 0.8420 & 0.6537 & 0.9090 & 0.9188 & 0.7380\\   
			& HyperIQA~\cite{hyperiqa} &0.8646 &0.8684 &0.6848 & 0.9168 & 0.9269 & 0.7497\\\midrule        
			\multirow{3}{*}{\shortstack{ Conventional \\NR \\ DQA}} & FADE~\cite{choi2015referenceless}  & 0.2502&0.1845 &0.1712 & 0.7283 & 0.6659 & 0.5319  \\
			& HazDesNet~\cite{zhang2020hazdesnet} &0.3175 &0.2999 &0.2158 & 0.6848 & 0.6875 & 0.4847 \\
			& BDQM~\cite{lv2023blind}& 0.7733 & 0.7858 & 0.5818 & 0.6465 & 0.5938 & 0.4738\\\midrule
			\multirow{3}{*}{\shortstack{CLIP-based \\ NR \\ DQA}}& CLIPIQA~\cite{wang2023exploring} & 0.3630 & 0.3545 & 0.2467 & 0.3142 & 0.3161 & 0.2168\\
			& CLIPIQA$^{+}$ ~\cite{wang2023exploring} & 0.8392 & 0.8510 & 0.6523 & 0.9111 & 0.9315 & 0.7394\\         
			& CLIP-DQA (Ours) & \textbf{0.9179} & \textbf{0.9232 }& \textbf{0.7568} & \textbf{0.9227} &\textbf{0.9376}&\textbf{0.7589}\\    
			\bottomrule
		\end{tabular}
            \begin{tablenotes}
                \footnotesize
                \item The performance of FR metrics on the DHQ dataset is not reported due to a lack of reference images.
              \end{tablenotes}
            \end{threeparttable}       
	}

	\label{tab:single}
\end{table}

\subsection{Performance Comparisons}
To validate the effectiveness of the proposed method, we compare it with 13 representative methods, which include three full-reference general image quality assessment (FR GIQA) methods, two FR DQA metrics, three no-reference general image quality assessment (NR GIQA) methods, three conventional NR DQA methods, and two CLIP-based NR DQA methods. All NR methods, except for FADE~\cite{choi2015referenceless} and HazDetNet~\cite{zhang2020hazdesnet}, are retrained with the same settings as ours. The results are reported in Table~\ref{tab:single}. From this table, we have the following observations. First, five FR IQA methods fail to produce results on the DHQ dataset due to the lack of haze-free images, and underperform on the exBeDDE dataset. Second, three NR GIQA methods exhibit competitive performance against three NR DQA methods on the two datasets. This is mainly because FADE and HazDesNet are designed for haze density prediction rather than quality evaluation, while BDQM is limited by its shallow network architecture. Third, the CLIP-based NR DQA method, CLIPIQA$^{+}$, is comparable to HyperIQA, which achieves the second-best performance on DHQ and exBeDDE datasets. This indicates the promising potential of CLIP models for NR DQA. Fourth, our proposed method, \textit{i.e.,} CLIP-DQA, outperforms HyperIQA by a noticeable margin, especially on the DHQ dataset. This confirms the effectiveness of our proposed method.

\begin{table}[tbp] 
	\centering
	\caption{ Ablation Study on each component of the proposed method.}
	\scalebox{0.85}{
		\begin{tabular}{l|c|c|c|c|c}
			\toprule
			\multicolumn{2}{l|}{\diagbox{Components}{Method}} & $M_1$ & $M_2$ & $M_3$ &  Ours \\	\midrule
			\multicolumn{2}{l|}{Handcrafted Text Prompts} &  \checkmark & & &\\
			\multicolumn{2}{l|}{Textual Prompt Tuning} &  & \checkmark& \checkmark& \checkmark\\ 
			\multicolumn{2}{l|}{Visual Prompt Tuning} & & & \checkmark & \checkmark\\ 		
			\multicolumn{2}{l|}{Global Information}  &  & &  & \checkmark \\ 		
			\multicolumn{2}{l|}{Local Information} &  \checkmark  & \checkmark  & \checkmark  & \checkmark \\ 		
			\midrule 
			\multirow{3}{*}{DHQ} & SRCC & 0.3630 & 0.8392 &  0.8925 & \textbf{0.9179}  \\ 		
			& PLCC & 0.3545& 0.8510  & 0.9002  & \textbf{0.9232}  \\ 		
			& KRCC & 0.2467& 0.6523 & 0.7200 &\textbf{0.7568}  \\ 	
			\midrule 
			\multirow{3}{*}{exBeDDE} & SRCC & 0.3142 & 0.8984  & 0.9111 & \textbf{0.9227}  \\ 		
			& PLCC & 0.3161& 0.9219 & 0.9315 & \textbf{0.9376}  \\ 		
			& KRCC & 0.2168  & 0.7182  & 0.7394  &\textbf{0.7589}  \\ 
			\bottomrule
	\end{tabular}}
	\label{tab:2}
\end{table}

\subsection{Ablation Study}

\noindent\textbf{Effectiveness of Textual Prompt Tuning.} We explore two variants of our proposed method, \textit{i.e.,} $M_1$ and $M_2$. Both methods disable visual prompt tuning and conduct the quality evaluation based on local information of dehazed images. However, $M_1$ uses handcrafted text prompts, while $M_2$ employs textual prompt tuning. The comparison results of these two methods are presented in Table~\ref{tab:2}. As we can see, $M_2$ significantly outperforms $M_1$. For example, on the DHQ dataset, the SRCC of $M_2$ is more than double the SRCC of $M_1$. Therefore, we can draw the conclusion that textual prompt tuning is effective.

\noindent\textbf{Effectiveness of Visual Prompt Tuning.} We explore another variant of our proposed method, i.e., $M_3$.  Compared to $M_2$ with only textual prompt tuning, $M_3$ turns on both textual and visual prompt tuning. As we can see, $M_3$ shows clear advantages over $M_2$ in all three evaluation metrics, especially on the DHQ dataset. As a result, we can conclude that visual prompt tuning is useful.

\noindent\textbf{Effectives of Hierarchical Information Perception.} Compared to $M_3$, our proposed method exploits both local and global information of dehazed images for quality prediction. As shown in Table~\ref{tab:2}, the proposed method achieves better performance than $M_3$ on both datasets. This confirms the efficacy of hierarchical information perception.

\begin{figure}[t]
\centering
\includegraphics[width=0.95\linewidth]{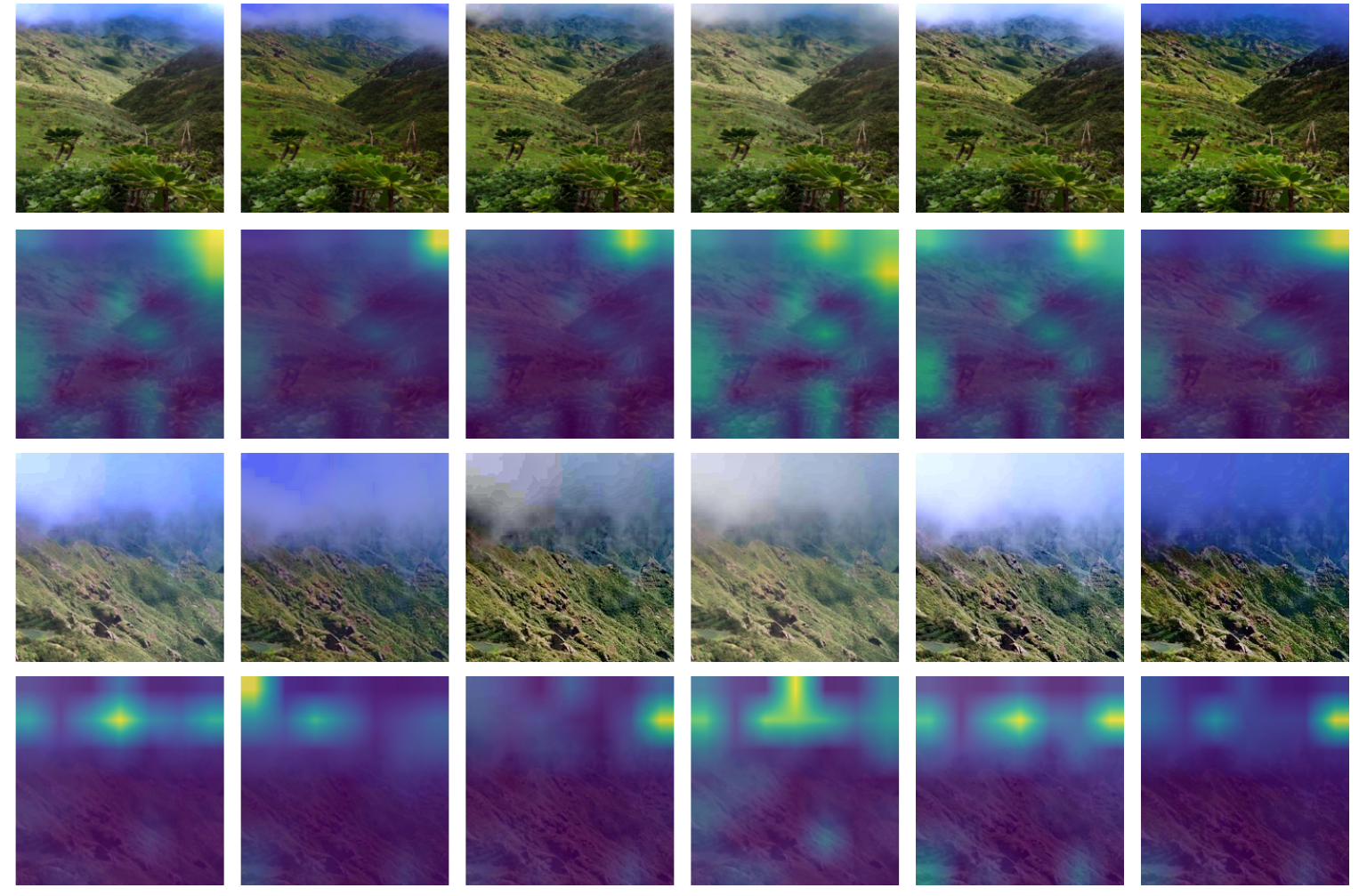}
	\caption{Illustration of the average attention map for the last visual transformer layer. The first and third rows are resized versions and patches of six dehazed images, respectively. The second and fourth rows visualize attention on resized images and patches, respectively. Each column corresponds to a dehazed image. }
	\label{fig:att}
\end{figure}

\subsection{Case Study}
To understand how the proposed method perceives dehazed images, we visualize the average attention map of the last visual transformer layer. In the average attention map, spatial regions with brighter colors contribute more to quality predictions. According to Fig.~\ref{fig:att}, we have the following findings:

First, for the resized version of dehazed images, the proposed method tends to spread attention to the whole image. Second, for patches cropped from dehazed images, the proposed method mainly focuses on hazy regions. In summary, the proposed method employs global and local information on dehazed images for quality prediction. This is highly consistent with the human visual system.
\section{Conclusion}
In this paper, we present the first preliminary study on introducing CLIP to BDQA, dubbed CLIP-DQA. Motivated by characteristics of the human vision system, CLIP-DQA blindly evaluates dehazed images from global and local perspectives. Furthermore, CLIP-DQA employs learnable multi-modal prompts to tune CLIP for accurate quality prediction. Comprehensive experiments demonstrate that the proposed method achieves state-of-the-art performance.

\bibliography{sample-base}
\bibliographystyle{IEEEtran}
\end{document}